\documentclass{article}

\PassOptionsToPackage{numbers, compress}{natbib}
 \usepackage[preprint]{neurips_2026}

\makeatletter
\DeclareRobustCommand\onedot{\futurelet\@let@token\@onedot}
\def\@onedot{\ifx\@let@token.\else.\null\fi\xspace}
 
\def\ie{\emph{i.e}\onedot}

\usepackage[most]{tcolorbox}
\usepackage{xcolor}

\definecolor{noteblue}{RGB}{225,239,255}
\definecolor{noteblueframe}{RGB}{40,120,200}

\newcommand{\ours}{{GeRo}\xspace}

\usepackage[utf8]{inputenc} 
\usepackage[T1]{fontenc}    
\usepackage{hyperref}       
\usepackage{url}            
\usepackage{booktabs}       
\usepackage{amsfonts}       
\usepackage{nicefrac}       
\usepackage{microtype}      
\usepackage{xcolor}         
\usepackage{multirow}
\usepackage{amsmath} 
\usepackage{amssymb}
\usepackage{makecell}
\usepackage{pifont}
\usepackage{media9}
\usepackage{graphicx}
\usepackage{wrapfig}
\usepackage{animate}
\usepackage{stfloats}
\usepackage{xspace}
\usepackage{enumitem}
\usepackage{caption}
\usepackage{subcaption}
\usepackage{marvosym}

\makeatletter
\title{Generative Scenario Rollouts for End-to-End Autonomous Driving}

\author{
\textbf{Rajeev Yasarla$^{1\text{\Letter}}$\quad Deepti Hegde$^1$\quad Shizhong Han$^1$\quad Hsin-Pai Cheng$^1$ \quad  Yunxiao Shi$^1$} \\[1pt]
\textbf{Meysam Sadeghigooghari$^2$\quad Shweta Mahajan$^{1}$\thanks{Work done while employed at Qualcomm AI Research.}\quad\,  Apratim Bhattacharyya$^1$} \\[1pt]
\textbf{Litian Liu$^1$\quad  Risheek Garrepalli$^1$ \quad Thomas Svantesson$^2$\quad  Fatih Porikli$^1$} \\[1pt]
\textbf{Mohammad Ghavamzadeh$^1$\quad  Hong Cai$^{1\text{\Letter}}$} \\[5pt]
$^1$Qualcomm AI Research\thanks{Qualcomm AI Research is an initiative of Qualcomm Technologies, Inc.} \quad $^2$Qualcomm Technologies, Inc.\\
[3pt]
\small{\texttt{\{ryasarla, hongcai\}@qti.qualcomm.com}}
}

\begin{document}

\maketitle

\begin{abstract}

\vspace{-8pt}

Vision-Language-Action (VLA) models are emerging as highly effective planning models for end-to-end autonomous driving systems. However, current works mostly rely on imitation learning from sparse trajectory annotations and under-utilize their potential as generative models. We propose Generative Scenario Rollouts (\ours), a plug-and-play framework for VLA models that jointly performs planning and generation of language-grounded future traffic scenes through an autoregressive rollout strategy.
First, a VLA model is trained to encode ego vehicle and agent dynamics into latent tokens under supervision from planning, motion, and language tasks, facilitating text-aligned generation.
Next, \ours performs language-conditioned autoregressive generation. Given multi-view images, a scenario description, and ego-action questions, it generates future latent tokens and textual responses to guide long-horizon rollouts. A rollout-consistency loss stabilizes predictions using ground truth or pseudo-labels, mitigating drift and preserving text-action alignment. 
This design enables \ours to perform temporally consistent, language-grounded rollouts that support long-horizon reasoning and multi-agent planning. 
On Bench2Drive, \ours improves driving score and success rate by +15.7 and +26.2, respectively. 
By integrating reinforcement learning with generative rollouts, \ours achieves state-of-the-art closed-loop and open-loop performance, demonstrating strong zero-shot robustness. 
These results highlight the promise of generative, language-conditioned reasoning as a foundation for safer and more interpretable end-to-end autonomous driving.

\end{abstract}

\vspace{-10pt}

\vspace{-5pt}
\section{Introduction}
\label{sec:intro}

\vspace{-8pt}


Autonomous driving systems require robust navigation and planning capabilities in complex and dynamic environments. Recent approaches train end-to-end (E2E) motion planners that map raw sensor inputs to future ego-vehicle trajectories~\cite{jiang2023vad,weng2024drive,zheng2024genad,sun2024sparsedrive,hu2023planning,yasarla2025roca}. Within this paradigm, 
Vision-Language-Action (VLA) models use large language models (LLMs) to integrate linguistic context into motion planning, enabling language-informed driving decisions~\cite{wang2024omnidrive,renz2025simlingo,zhou2025autovla,tian2024drivevlm,fu2024drive,chen2024driving, fu2025orion}.

\begin{figure}[t!]
    \centering
    \includegraphics[width=0.95\linewidth]{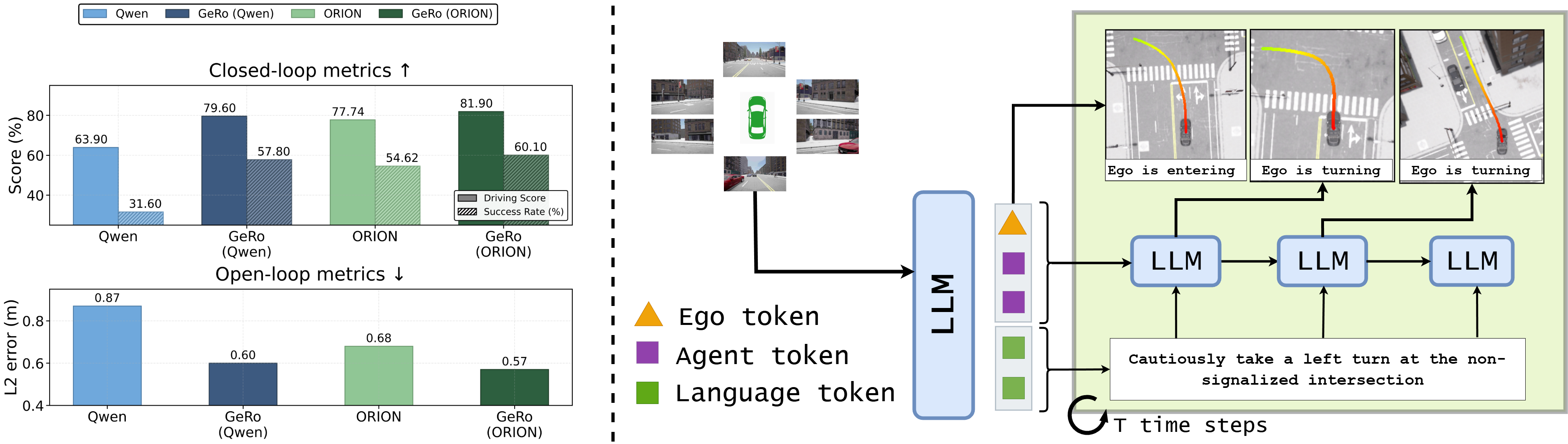}
    \vspace{-5pt}
    \caption{\small (\textit{Left}) Closed-loop metrics (Driving Score and Success Rate) and open-loop metric (Trajectory L2 error) for different models. \ours consistently improves both closed-loop and open-loop performance across baselines like Qwen2.5~\cite{Qwen2.5-VL} and ORION~\cite{fu2025orion}.
    (\textit{Right}) \ours scenario rollout pipeline: given scenario description and multi-view observations, \ours generates temporally consistent agent behaviors using a large language model (LLM) across time steps. }
    \label{fig:intro_overview}
    \vspace{-13pt}
\end{figure}

Despite this progress, current VLA models exhibit critical limitations and leave several directions underexplored. 
    (1) Sparse language-action supervision: Driving datasets \cite{sima2023drivelm,qian2024nuscenes,fu2025orion} often provide scene-level descriptions and Q\&A pairs but rarely include fine-grained actions tied to temporal phases of driving events, making models brittle in ambiguous or long-tail scenarios, e.g., distinguishing overtaking from merging,  localizing the ego vehicle’s intent within multi-step maneuvers \cite{fu2025orion}.
    (2) Underutilized generative capability: Existing VLA methods rely only on ground-truth trajectories for planning, overlooking the potential of autoregressive generation for scenario-level reasoning and exploration. 
    (3) Descriptive vs. procedural language: Current language supervision typically describes what is happening rather than how actions unfold, limiting the model’s ability to capture procedural nuances necessary for planning. 
    (4) Language-action misalignment: Many datasets generate instruction–action pairs after collecting expert driving data~\cite{renz2025simlingo}, so models often infer from visual cues only and ignore language, causing failures like “stop for red light” paired with acceleration.
To address these issues, we propose Generative Scenario Rollouts (\ours), a plug-and-play training framework that unifies VLA models with scenario generation for language-aligned action prediction and improved planning robustness (Figure~\ref{fig:intro_overview}). To the best of our knowledge, this is the first work to jointly perform scenario generation with motion prediction, planning, and visual question-answering. \ours has two phases:
\begin{itemize}[leftmargin=*, itemsep=0em, topsep=0em]
    \item \textbf{Pretraining:} In this stage, the VLA model is trained to encode ego and agent dynamics into a compact shared token space. The model is jointly supervised for planning, multi-agent motion prediction with auxiliary bounding-box cues, and visual question answering. This stage is crucial for binding linguistic and behavioral representations, enabling text-aligned generation beyond paired annotations, thus reducing language-action misalignment.  
    \item \textbf{Language-Conditioned Scenario Rollout:} In this phase, the model autoregressively predicts future latent tokens and ego-aligned action descriptions based on past latent tokens. The tokens are fed back to the VLA model to guide long-horizon reasoning. Predictions are stabilized using a consistency loss that mitigates drift and enforces text-action consistency.    Specifically, we enforce temporal consistency by aligning the latent distribution of rollout predictions with the pretrained latent distribution for future time steps using KL-Divergence. When ground-truth data is available, imitation learning is applied by supervising the model with corresponding waypoints and motion trajectories to enforce scenario generation. However, relying solely on imitation learning can lead to suboptimal performance; therefore, we introduce reinforcement learning based feedback with GRPO \cite{shao2024deepseekmath} for the  scenario rollouts. This incorporates rewards relating to collision avoidance, time-to-collision (TTC), and language-aligned scenario generation using question–answer pairs relating to the action of the ego-vehicle. 
\end{itemize}


Our contributions are threefold:
\begin{enumerate}[leftmargin=*, itemsep=0em, topsep=0em]
    \item We introduce \textbf{\ours}, a generative scenario rollout framework that integrates autoregressive scenario generation into VLA models. This design couples temporal reasoning with linguistic context, enabling scenario rollouts that predict future motions of traffic agents and ego actions for robust planning.
    \item We propose a novel set of reward functions for GRPO-based reinforcement learning that jointly optimize trajectory accuracy and semantic alignment with language descriptions during scenario rollouts. These rewards incorporate safety-critical metrics such as collision avoidance and time-to-collision, ensuring high-fidelity and interpretable planning behavior.
    \item \textbf{\ours} introduces an interactive Visual Question Answering (VQA) component that grounds ego-vehicle intent in natural language, allowing the model to answer scenario-specific questions during rollouts. This enhances interpretability and provides a mechanism for language-guided reasoning in complex driving environments.
\end{enumerate}

Through extensive closed-loop and open-loop experiments, we show that \ours significantly outperforms VLA baselines \cite{fu2025orion} and other E2E motion planners \cite{jiang2023vad,hu2023planning,jia2025drivetransformer}, particularly in corner cases and zero-shot planning, highlighting the importance of scenario-level reasoning and language-action alignment for E2E autonomous driving.

\section{Related Works}
\label{sec:related}

\vspace{-7pt}
\subsection{VLA for Autonomous Driving}
\vspace{-7pt}
While conventional E2E models like VAD~\cite{jiang2023vad} and UniAD~\cite{hu2023planning} offer fast inference and streamlined integration of perception, prediction, and planning, they often struggle with generalization in long-tail scenarios. Recent LLM-based approaches~\cite{touvron2023llama, dubey2024llama, fu2025orion,wang2024omnidrive,renz2025simlingo,zhou2025autovla,tian2024drivevlm,fu2024drive,chen2024driving,yasarla2025roca,renz2025simlingo,hegde2025distilling} integrate world-model reasoning with language, improving robustness in long-tail scenarios. GPT-Driver~\cite{mao2023gpt} reformulates planning as text generation, converting observations into language prompts and leveraging chain-of-thought reasoning~\cite{wei2022chain}. EMMA~\cite{hwang2024emma} enables models like Gemini~\cite{chen2024driving} to predict discrete textual plans with strong open-loop performance. Hybrid dual-system architectures combine fast E2E models with slower, reasoning-capable VLMs. DriveVLM~\cite{tian2024drivevlm} employs a dual system where a VLM predicts low-frequency trajectories refined by a conventional planner. Senna~\cite{fu2024drive} introduces meta-actions to guide VAD for semantically aligned motion trajectories. ORION~\cite{fu2025orion} proposes a unified E2E framework that aligns VLM reasoning with trajectory prediction for closed-loop scenarios. SimLingo~\cite{renz2025simlingo} further improves language-scene alignment, achieving competitive accuracy on Bench2Drive, leveraging additional, higher-quality training data.

\vspace{-5pt}
\subsection{Driving Scenario Generation}
\vspace{-7pt}
Generative models, particularly diffusion-based ones, have shown promise in modeling the stochastic nature of driving behavior and generating diverse, multimodal trajectories. DiffusionDrive~\cite{liao2025diffusiondrive} introduces a truncated diffusion process with anchor priors, enabling real-time multimodal planning. GoalFlow~\cite{xing2025goalflow} applies goal-point guidance with flow matching for high-quality trajectory generation, while DiffusionPlanner~\cite{zheng2025diffusion} redefines planning as future trajectory generation, jointly producing plans and motion forecasts. InfGen proposes to do arbitrary resolution image generation in the latent space with diffusion models~\cite{han2025infgen}. In the VLA context, ORION extends its framework with a VAE-based generative planner that translates LLM outputs into trajectory distributions~\cite{fu2025orion}. DiVLA~\cite{liu2023mtd} combines autoregressive reasoning with diffusion policy learning, introducing a reasoning injection module that enhances interpretability and generalization. These generative VLA models establish a foundation for explainable, multimodal driving policies, motivating our generative rollout approach.

\begin{figure*}[t]
    \centering
    \includegraphics[width=0.9\textwidth]{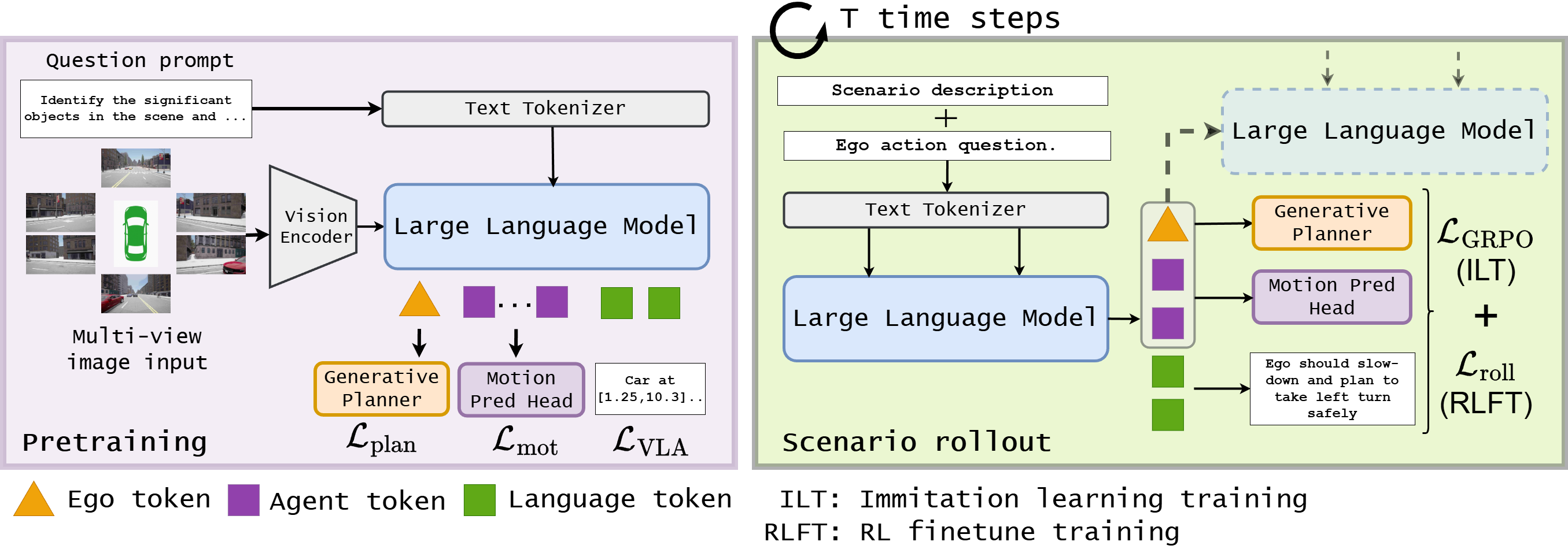}
    \vspace{-3pt}
    \caption{\small \textbf{Overview of \ours's two stage framework.} 
    (\textit{Left}) \ours~pretraining stage of the Vision–Language–Action (VLA) model. Multi-view images and scenario prompts are encoded into visual and text tokens and projected by the LLM head to compute ego and agent tokens. These tokens are passed to the \textit{generative planner} and the \textit{motion prediction head}. The planning, and motion and language prediction tasks are supervised by their respective loss functions (\( \mathcal{L}_{\text{plan}}, \mathcal{L}_{\text{mot}}, \mathcal{L}_{\text{VLA}}\)).  
    (\textit{Right}) \ours's autoregressive rollout stage. Given a scenario description and ego-action questions, \ours performs scenario rollout to predict future tokens for \( T \) steps, decoding them into ego waypoints, multi-agent trajectories, and language outputs. Rollouts are optimized with consistency loss (\( \mathcal{L}_{\text{roll}} \)) and reinforcement feedback (\( \mathcal{L}_{GRPO} \)) for robust scenario grounding.}
    \label{fig:GeSA_overview}
    \vspace{-12pt}
\end{figure*}

\vspace{-5pt}
\subsection{Reinforcement Learning for Driving}
\vspace{-7pt}
Reinforcement learning (RL) has emerged as a powerful paradigm for improving policy generalization and robustness in autonomous driving. Unlike imitation learning, which suffers from covariate shift~\cite{youssef2024covariateshift}, RL enables agents to learn from interaction and adapt to novel scenarios. RAD~\cite{gao2025rad} trains E2E agents in photorealistic 3DGS environments, demonstrating improved generalization. CarPlanner~\cite{zhang2025carplanner} introduces an autoregressive policy that generates consistent multimodal trajectories, outperforming imitation-based baselines through expert-guided reward shaping. Recent efforts also explore synergy between RL and VLMs: GenDrive~\cite{huang2025gen} combines diffusion models with RL and reward modeling, while~\cite{jiang2025alphadrive, liu2025reinforced,li2025finetuning} incorporate GRPO-based training to enhance policy stability and generalization. By integrating VLMs into the RL loop, these approaches enable semantic grounding and improve decision-making in complex, long-tail scenarios.

\vspace{-6pt}
\section{Methodology}
\vspace{-8pt}
We present \ours, a generative scenario rollout framework for VLA models that unifies planning and autoregressive scenario generation. \ours  is shown in Figure~\ref{fig:GeSA_overview}. In \ours, rollout operates in a token space that is learned in a pretraining stage, where ego and agent states are represented as compact latent tokens. Given latent ego-tokens and agent tokens along with a high-level scenario description and a question regarding the actions of the ego-vehicle, the \ours framework uses a VLA model to autoregressively generate future scenarios by predicting their latent ego and agent tokens. The model is also jointly trained to perform planning, motion prediction, and visual question-answering. These tasks are supervised under GRPO \cite{shao2024deepseekmath} with a  series of specially designed reward functions. During rollout training, GRPO \cite{shao2024deepseekmath} complements the supervised losses with safety and semantic rewards. An overview of the rollout mechanism can be seen in  Figure~\ref{fig:GeSA_rollout}.

\vspace{-7pt}
\subsection{Problem Formulation}
\vspace{-7pt}
 Consider a driving scene at time $t$ consisting of an ego vehicle and a set of \(N_t\) traffic participants \(\mathcal{A}_t\). Given a sequence of multi-view images, a scene description, and information about the state of the ego-vehicle \(x_t^e\) and the state of the other traffic participants \(x_t^{(a_i)}\), the objective of \ours is to jointly predict (i) an ego trajectory \(\hat{\tau}_{t}^e\) that supports safe and efficient driving, (ii) future motions of surrounding agents \(\{\hat{\tau}_{t}^{(a_i)}\}_{i=1}^{N_t}\),\footnote{For notational simplicity, we drop the agent indices when appropriate. For example, we refer to \(\{\hat{\tau}_{t}^{(a_i)}\}_{i=1}^{N_t}\)  as \(\{\hat{\tau}_{t}^{(a_i)}\}\), and we refer to the set of ego and agent trajectories, \(\{\hat{\tau}^e,\{\hat{\tau}_{t}^{(a_i)}\}_{i=1}^{N_t}\}\) as \(\{\hat{\tau}_{t}\}\), when it is clear from the context.} and (iii) a scenario-based ego-action answer to question $q$. The ego state comprises of pose, velocity, and acceleration. The state of agent $a_i$ consists of the agent's class, bounding box, motion trajectory, velocity, and acceleration.  

\vspace{-7pt}
\subsection{Architecture}

\vspace{-7pt}
\ours provides a general plug-and-play framework for Vision-Language-Action models. Thus, the architecture follows several recent end-to-end planners that leverage multi-modal LLMs for motion planning~\cite{fu2025orion,wang2024omnidrive,tian2024tokenize}. The vision encoder and the text tokenizer project image and text inputs to a common token embedding space. The central component is the LLM, which projects multi-modal tokens into a series of compact latent representations of the ego-vehicle and the surrounding agents. The LLM also predicts a scene description and answers questions related to the ego-vehicle behavior. Lastly, the framework contains a variational autoencoder (VAE) planning head and a motion prediction head. 

\begin{figure*}[t!]
    \centering
    \includegraphics[width=0.9\linewidth]{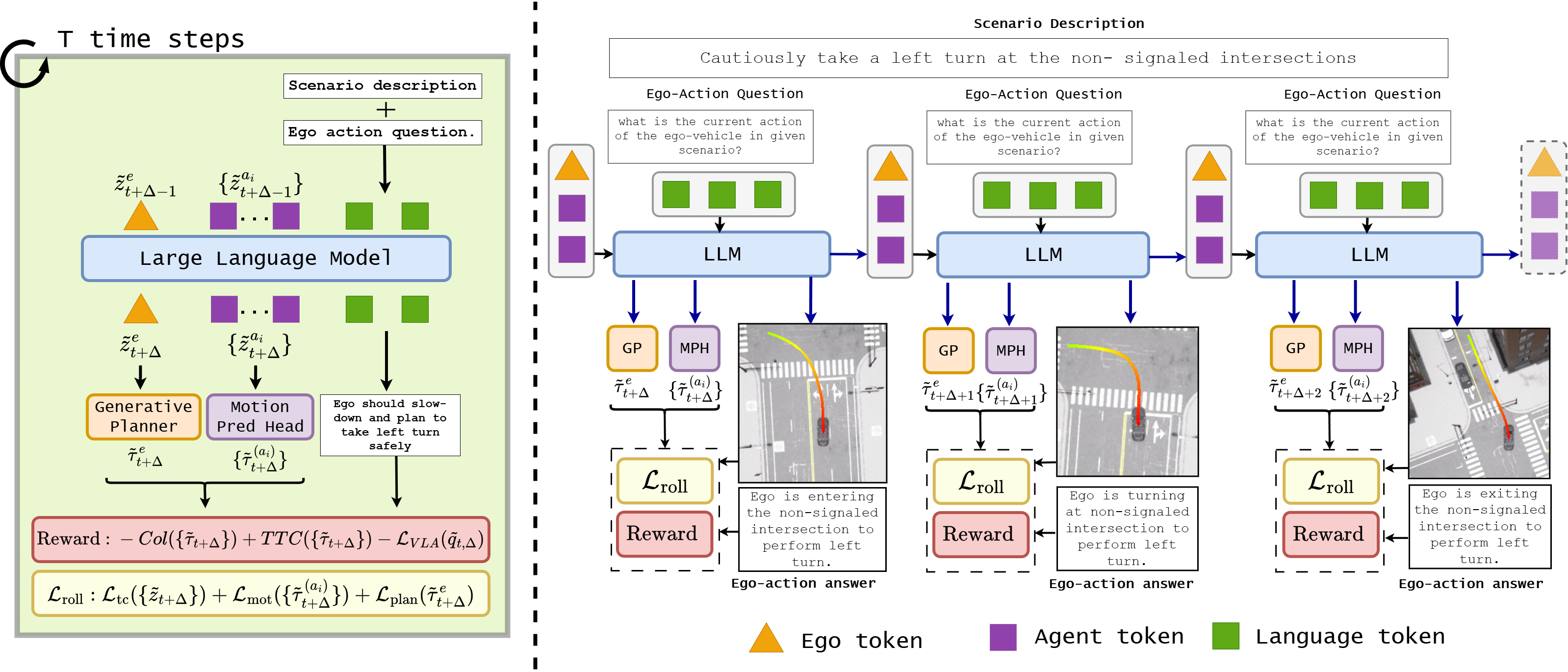}
    \vspace{-8pt}
    \caption{\small \textbf{Autoregressive scenario rollout in \ours.} 
    At time \( t \), \ours consumes ego and agent tokens along with scenario description and ego-action questions, and predicts next-step tokens using the LLM head. These tokens are decoded into ego trajectories, multi-agent trajectories, and language answers. The process repeats for \( T \) steps, with updated scenario descriptions and ego-action questions at each rollout step, enabling consistent rollouts aligned with text and environment dynamics. Rollouts are optimized using the consistency loss \( \mathcal{L}_{\text{roll}} \) and the reinforcement feedback loss \( \mathcal{L}_{\text{GRPO}} \) for robust planning and reasoning.}
    \label{fig:GeSA_rollout}
    \vspace{-12pt}
\end{figure*}

\vspace{-5pt}
\subsection{Pretraining}\label{sec:pretrain_gesa}
\vspace{-5pt}
The objective of the pretraining stage is to achieve a good initialization of action tokens for stable scenario generation in the token space. To this end, we first pretrain the VLA model to learn a compact tokenization of ego and agent dynamics. In this stage, we supervise three heads: (i) a planning head, which predicts future ego trajectory, (ii) the motion head, which predicts future multi-agent trajectories, and (iii) the large language model, which is trained to predict scene descriptions and perform visual question answering. The model is also trained to predict the detection bounding box of the agents along with their corresponding trajectories, \(\{\hat{\tau}_{t}^{(a_i)}, \hat{\phi}_{t}^{(a_i)}\}_{i=1}^{N_t}\) to accelerate convergence of trajectory prediction in the motion head. This pretraining stage teaches the model to encode and decode latent tokens for both ego and agents that remain consistent and representative of the scenario.

\vspace{-12pt}
\paragraph{Pretraining loss functions.}
The planning head is supervised by using two losses: an L1 loss for waypoint regression and a collision loss. The motion prediction head is supervised using (i) focal loss \cite{lin2017focal} for classification of agents' types, (ii) L1 loss for trajectory prediction, and (iii) L1 loss for 3D bounding box detection. The LLM is trained with a cross-entropy loss for language prediction, following \cite{Qwen2.5-VL}. Let \(\mathcal{L}_{\text{plan}}\), \(\mathcal{L}_{\text{mot}}\), and \(\mathcal{L}_{\text{VLA}}\) denote these respective terms; the overall pretraining loss is,
\begin{equation}\label{eqn:pretraining}
\mathcal{L}_{\text{pre}} = \mathcal{L}_{\text{plan}} + \mathcal{L}_{\text{mot}} + \mathcal{L}_{\text{VLA}}.
\end{equation}
\subsection{Generative Scenario Rollout}\label{sec:gen_scene}
\vspace{-5pt}
Generative scenario rollout training comprises two sequential phases. First, imitation learning with \(\mathcal{L}_{\mathrm{roll}}\) learns temporally consistent latent futures and accurate ego and agent trajectories from logged driving data. We then apply GRPO-based reinforcement learning with \(\mathcal{L}_{\mathrm{GRPO}}\) to refine the model using safety feedback from generated rollouts.
\paragraph{Scenario Rollout.}
Given a sequence of multi-view images up to time $t$, \ours first computes the latent ego token \( z_t^e \), the latent agent tokens \( \{ z_t^{(a_i)} \} \), and scenario description \( s_t \), as shown in Figure~\ref{fig:GeSA_overview}. Next, \ours performs autoregressive rollouts in the token space for \( T \) total steps as shown in Figure~\ref{fig:GeSA_rollout}. Conditioned on latent tokens \(\{ z_t \}\), multi-view images at time $t$, a scenario description \( s \), and the ego-action question $q_{t, 1} \in Q$, the generator predicts the next-step tokens \( \{ \tilde{z}_{t+1} \} \) and the responses to the ego-action questions. These tokens are decoded into the ego trajectory, agent motion trajectories, and language outputs. The predicted tokens are fed back to the model along with a scenario description \( s \) and the next ego-action question \( q_{t,\Delta+1} \), where \( \Delta \in [0, T-1] \). This enables the model to generate the subsequent tokens \( \{ \tilde{z}_{t+\Delta + 1} \} \), resulting in a rollout from $t$ to  $t+T$.
\vspace{-5pt}
\paragraph{{Language Supervision.}} 
A central goal of \ours is to generate scenario rollouts of action tokens while grounding them in language, enabling the model not only to produce semantically coherent sequences of future tokens but also to reason about the context. Rollout is guided by a scene description $s$ provided by the LLM and a question $q$ related to the action of the ego vehicle. Given each image sequence up to time $t$, its high-level scenario category, critical object information, and language annotations, the LLM is trained to then generate a rich scenario description \( s \) and corresponding ego-action questions \( Q \) that serve as language guidance for the autoregressive rollout.  
\vspace{-5pt}
\paragraph{Enforcing Rollout Consistency.}
During scenario rollout, temporal consistency is enforced through two complementary mechanisms. First, the latent distribution of rollout predictions are aligned with the pretrained latent distribution for future time steps using KL-Divergence. Second, the model is supervised using imitation learning when ground truth labels are available.

When ground truth trajectory labels are unavailable, they can be constructed from latent tokens ${z_{t+\Delta}}$ predicted by the pretrained VLA model and conditioned on the input image sequence with additional future frames. This setup promotes temporal consistency across extended rollouts and helps mitigate compounding errors commonly observed in autoregressive generation.

To stabilize training over long horizons, we apply step-wise temporal consistency supervision $\mathcal{L}_{tc}$ using pretraining heads under two regimes:

\begin{itemize}[leftmargin=*, itemsep=0em, topsep=0em]
    \item \textbf{Ground-Truth Supervision:} When ground-truth data is available, we supervise the model using true states \(\{x_{t+\Delta}\}\) and trajectories \(\{\tau_{t+\Delta}\}\), providing accurate targets for each rollout step. 
    \item \textbf{Model-Based Supervision:} In the absence of ground-truth, we rely on tokens \(\{z_{t+\Delta}\}\) predicted by the pre-trained VLA model (refer to Section~\ref{sec:pretrain_gesa}) using corresponding input image sequence as reference targets. These predictions serve as pseudo-labels to guide \ours during unsupervised rollout steps. The token-consistency loss is defined as: \(\mathcal{L}_{\text{tc}}\big(\{\tilde{z}_{t+\Delta}\}\big) = \text{KL}\big(\{\tilde{z}_{t+\Delta}\} \,\|\, \{z_{t+\Delta}\}\big),\)
    where \(\text{KL}(\cdot\|\cdot)\) denotes the Kullback--Leibler divergence. This formulation enforces temporal consistency in scenario generation by aligning the rollout predictions and the latent distribution of the pre-trained model.
\end{itemize}

This temporal consistency loss $\mathcal{L}_{\text{tc}}$ encourages alignment between the predicted and reference sequences—whether derived from ground-truth or VLA-generated pseudo-labels. As a result, this rollout loss reduces drift and improves the fidelity of long-horizon planning. The rollout loss is formulated by integrating both planning and motion prediction components as follows:
\setlength{\abovedisplayskip}{2pt}
\setlength{\belowdisplayskip}{2pt}
\begin{equation}
\label{eqn:rollout_loss}
\mathcal{L}_{\text{roll}} =
\sum_{\Delta=1}^{T} \Big[
    \mathcal{L}_{\text{tc}}(\{\tilde{z}_{t+\Delta}\})
  + \mathcal{L}_{\text{plan}}(\{\tilde{\tau}_{t+\Delta}\})
  + \mathcal{L}_{\text{mot}}(\{\tilde{\tau}_{t+\Delta}\})
\Big]
\end{equation}
\vspace{-15pt}
\paragraph{RL with GRPO.}

After the imitation learning phase, we train our model using Group Relative Policy Optimization (GRPO)~\cite{shao2024deepseekmath} with safety and semantic feedback. Given the multi-modal scene features \(x_t\), we generate a group \(\{\tau_g\}_{g=1}^{G}\) of \(G=8\) rollouts. Each rollout \(\tau_g\) contains ego waypoints $\{\tilde{\tau}_{g,t+\Delta}^e\}_{\Delta=1}^T$ and surrounding-agent trajectories $\{\tilde{\tau}^{(a_i)}_{t+\Delta}\}$, where the ego waypoints are sampled from the generative planner and agent motion trajectories predicted through the motion prediction head. 
Each rollout receives a scalar reward aggregating collision (\(\operatorname{Coll}\)), time-to-collision (TTC), and language-consistency terms over the rollout steps, i.e.,
\begin{equation}
R_g = \sum_{\Delta=1}^{T}\Big[
 -\operatorname{Coll}(\{\tilde{\tau}_{g,t+\Delta}\})
 +\operatorname{TTC}(\{\tilde{\tau}_{g,t+\Delta}\})
 -\mathcal{L}_{\text{VLA}}(\tilde{y}_{g,t+\Delta})
\Big],
\end{equation}
where \(\tilde{y}_{g,t+\Delta}\) denotes the generated language response, which is scored against its reference using \(\mathcal{L}_{\text{VLA}}\). We compute the group baseline and normalized advantage as
\begin{equation*}
b = \frac{1}{G}\sum_{g=1}^{G}R_g, \qquad\quad \sigma_R = \operatorname{std}\{R_g\}_{g=1}^G, \qquad\quad A_g = \frac{R_g-b}{\sigma_R+\zeta},
\end{equation*}
%
where $\zeta$ is a small positive number, and optimize the policy $\pi_\theta(\cdot|x)$ using the GRPO objective  
%
\begin{equation}
\label{eqn:grpo-loss}
\mathcal{L}_{\text{GRPO}} = - \frac{1}{G}\sum_{g=1}^G \min\Big(r_g(\theta)A_g,\;
\operatorname{clip}\big(r_g(\theta),1-\epsilon,1+\epsilon\big)A_g\Big)
\end{equation}
where $r_g(\theta)=\frac{\pi_\theta\big(\{\tilde{\tau}_{g,t+\Delta}\} \mid x\big)}{\pi_{\theta_{\text{old}}}\big(\{\tilde{\tau}_{g,t+\Delta}\} \mid x\big)}$ and the clip parameter is set to \(\epsilon=0.1\). We use GRPO with batch size 256 and mini-batches of size 32. 


\subsection{Overall Supervision}
\vspace{-5pt}
The loss functions used in the two stages of \ours are as follows: 
\begin{itemize}[leftmargin=*, itemsep=0em, topsep=0em]
    \item \textbf{Stage-1: Pretraining.} The VLA model is trained to learn a representative action token space under the pretraining objective \(\mathcal{L}_{\text{pre}}\), summarized in Eq.~\ref{eqn:pretraining}.
    \item \textbf{Stage-2: Scenario Rollout Generation.}
The VLA is optimized in two sequential phases: imitation learning using
\(\mathcal{L}=\mathcal{L}_{\mathrm{roll}}\) (Eq.~\ref{eqn:rollout_loss}), followed by
GRPO-based reinforcement learning using
\(\mathcal{L}=\mathcal{L}_{\mathrm{GRPO}}\) (Eq.~\ref{eqn:grpo-loss}).
\end{itemize}
Here, \(\mathcal{L}_{\text{roll}}\) enforces consistency across autoregressive rollout, while \(\mathcal{L}_{\text{GRPO}}\) incorporates GRPO-based reinforcement feedback for robust scenario grounding. 

\begin{table*}[t!]
\centering
\caption{\small Comparison of closed-loop and open-loop performance of end-to-end autonomous driving methods on the Bench2Drive base set. Avg.~L2 denotes the average trajectory error over 2 seconds at 2~Hz, following UniAD~\cite{hu2023planning}. Abbreviations: NC = Navigation Command, TP = Target Point, DS = Driving Score, SR = Success Rate, M = Map, B = Bounding boxes, D = Depth, Mo = Motion prediction, O = Occupancy, S = Segmentation, E = Expert distillation. $\ddagger$ indicates models trained with geometric path waypoint supervision and evaluated using the PID controller from SimLingo.}
\vspace{-0.6em}
\label{tab:b2d_results}
\resizebox{0.95\textwidth}{!}{
\begin{tabular}{lccccccc}
\toprule
\multirow{2}{*}{Method} & \multirow{2}{*}{Condition} & \multirow{2}{*}{Aux. Sup.} & \multicolumn{4}{c}{Closed-loop Metric} & \multicolumn{1}{c}{Open-loop Metric} \\
\cmidrule(lr){4-7} \cmidrule(lr){8-8}
& & & DS$\uparrow$ & SR(\%)$\uparrow$ & Efficiency$\uparrow$ & Comfortness$\uparrow$ & Avg. L2 $\downarrow$ \\
\midrule
TCP-traj~\cite{wu2022trajectory} & TP & E+Speed & 59.90 & 30.00 & 76.54 & 18.08 & 1.70 \\
ThinkTwice~\cite{jia2023thinktwice} & TP & E+D+S+M & 62.44 & 31.23 & 69.33 & 16.22 & 0.95 \\
DriveAdapter~\cite{jia2023driveadapter}  & TP & E+S & 64.22 & 33.08 & 70.22 & 16.01 & 1.01 \\
SimLingo~\cite{renz2025simlingo}$\ddagger$ & TP & - & 85.07 & 67.27 & \textbf{259.23} & 33.67 & - \\ 
iPad~\cite{guo2025ipad}  & TP & - & 65.02 & 35.91 & 161.31 & 28.21 & 0.97 \\
RAP-ResNet~\cite{feng2025rap} & TP & 3D Raster & 66.42 & 37.27 & 165.47 & 23.63 & - \\
\hline
UniAD-Base~\cite{hu2023planning}  & NC & M+B+Mo+O & 45.81 & 16.36 & 129.21 & 43.58 & 0.73 \\
VAD \cite{jiang2023vad}   & NC & M+B+Mo & 42.35 & 15.00 & 157.94 & 46.01 & 0.91 \\
GenAD~\cite{zheng2024genad}   & NC & M+B+Mo & 44.81 & 15.90 & - & - & - \\
MomAD~\cite{song2025don}  & NC & M+B+Mo & 44.54 & 16.71 & 170.21 & \textbf{48.63} & 0.87 \\
DriveTransformer-Large~\cite{jia2025drivetransformer}   & NC & M+B+Mo & 63.46 & 35.01 & 100.64 & 20.78 & 0.62 \\
ReCogDrive~\cite{li2025recogdrive}  & NC & - & 71.36 & 45.45 & 138.18 & 17.45 & -\\ 
HiP-AD~\cite{tang2025hip}$\ddagger$  & NC & M+B+Mo & 86.77 & 69.09 & 203.12 & 19.36 & 0.69\\ 
Qwen2.5~\cite{Qwen2.5-VL} & NC & B+Mo & 63.90 & 31.60 & 119.34 & 10.11 & 0.87 \\ 
ORION~\cite{fu2025orion}  & NC & M+B+Mo & 77.74 & 54.62 & 151.48 & 17.38 & 0.68 \\ 
ORION~(Qwen2.5)~\cite{fu2025orion} & NC & M+B+Mo & 76.88 & 53.90 & 146.16 & 24.75 & 0.72\\ 
\hline
\rule{0pt}{1.2em}\ours (Qwen2.5) (ours)  & NC & B+Mo & 79.60 & 57.80 & 159.66 & 44.75 & 0.60 \\
\rule{0pt}{1.2em}\ours (ORION) (ours)  & NC & M+B+Mo & 81.90 & 60.10 & 176.51 & 40.22 & 0.57 \\
\rule{0pt}{1.2em}\ours (ORION) (ours)$\ddagger$  & TP & M+B+Mo & \textbf{87.16} & \textbf{68.76} & 182.94 & 38.27 & \textbf{0.56} \\
\bottomrule

\end{tabular}
}
\vspace{-10pt}
\end{table*}

\vspace{-4pt}
\section{Experimental setup}
\label{sec:experiments}

\vspace{-6pt}
\subsection{Dataset and Evaluation Metrics}
\vspace{-7pt}
\paragraph{Dataset.} \ours is trained and evaluated on the Bench2Drive~\cite{jia2024bench2drive} dataset, a closed-loop benchmark based on CARLA \cite{dosovitskiy2017carla} for end-to-end autonomous driving. We use the official set of 1000 clips (950 for training, 50 for open-loop validation), each covering about 150 meters of driving in diverse traffic scenes. Closed-loop evaluation follows the official protocol on 220 routes across 44 interactive scenarios. For open-loop planning, we use the nuScenes~\cite{caesar2020nuscenes} planning benchmark, which contains 28{,}000 samples in a 22k/6k training/validation split.  We additionally evaluate Qwen and \ours~(Qwen) with camera-only inputs on the NAVSIM-v1 NavTest split~\cite{dauner2024navsim}, a data-driven non-reactive simulation benchmark based on real-world driving scenes.
\vspace{-10pt}
\paragraph{VQA Dataset.} 
Language annotations for the scene description and visual question answering tasks are obtained from the ChatB2D~\cite{fu2025orion} dataset in conjunction with Bench2Drive~\cite{jia2024bench2drive}, and the DriveLM-nuScenes~\cite{sima2023drivelm} dataset. We generate additional scenario descriptions and ego-action question-answer pairs for training \ours during scenario generation rollout (more details in appendix section \ref{sec:language_anno}).
\vspace{-10pt}
\paragraph{Metrics.} Bench2Drive closed-loop metrics include Driving Score (DS), Success Rate (SR), Efficiency, Comfort, and Multi-Ability. DS combines route completion and violation penalties; SR measures successful completion; Efficiency and Comfort assess speed and smoothness. Multi-Ability evaluates five advanced urban driving skills. Open-loop evaluation uses L2 trajectory error and collision rate, following ST-P3~\cite{hu2022st}. Open-loop nuScenes evaluation uses L2 trajectory error and collision rate with ST-P3-style metrics~\cite{hu2022st} under the PARA-Drive\cite{weng2024drive}/TOKEN\cite{pmlr-v270-tian25b} evaluation setting. For NAVSIM, we report No at-fault Collision (NC), Ego Progress (EP), and Predictive Driver Model Score (PDMS); higher is better for all three metrics.

\begin{table*}[t!]
\centering
\caption{\small Multi-ability performance Comparison for E2E-AD methods on Bench2Drive base set. NC = Navigation Command, TP = Target Point, M = Map, B = Bounding boxes, D = Depth, Mo = Motion prediction, O = Occupancy, S = Segmentation, E = Expert distillation. $\ddagger$ indicates models trained with geometric path waypoint supervision and evaluated using the PID controller from SimLingo.}
\vspace{-0.6em}
\label{tab:multi_ability}
\resizebox{\textwidth}{!}{
\begin{tabular}{lcccccccc}
\hline
\multirow{2}{*}{Method} & \multirow{2}{*}{Condition} & \multirow{2}{*}{Aux. Sup.} & \multicolumn{6}{c}{Ability (\%) $\uparrow$} \\
\cmidrule(lr){4-9}
  & & & Merging & Overtaking & Emergency Brake & Give Way & Traffic Sign & Mean \\
\hline
TCP-traj~\cite{wu2022trajectory}  & TP & E+Speed & 28.84 & 24.29 & 51.67 & 40 & 46.28 & 34.22 \\
ThinkTwice~\cite{jia2023thinktwice}  & TP & E+D+S+M & 27.38 & 18.42 & 35.82 & 50 & 54.23 & 37.17 \\
DriveAdapter~\cite{jia2023driveadapter}  & TP & E+S & 28.38 & 28.38 & 47.50 & 50 & 56.43 & 42.14 \\
SimLingo~\cite{renz2025simlingo}$\ddagger$  & TP & - & \textbf{54.01} & 57.04 & 88.33 & 53 & \textbf{82.45} & 67.03 \\
\hline
UniAD-Base~\cite{hu2023planning}  & NC & M+B+Mo+O & 8.89 & 9.33 & 20 & 20 & 14.21 & 14.49 \\
VAD \cite{jiang2023vad}  & NC & M+B+Mo & 11.43 & 11.43 & 18.64 & 20 & 19.15 & 18.07 \\
DriveTransformer-L \cite{jia2025drivetransformer}  & NC & M+B+Mo & 17.57 & 35.00 & 48.36 & 40 & 52.10 & 38.60 \\

HiP-AD~\cite{tang2025hip}$\ddagger$  & NC & M+B+Mo & 50.00 & 84.44 & 83.33 & 40 & 50.45 & 65.98 \\
ReCogDrive~\cite{li2025recogdrive}  & NC & - & 29.73 & 20.00 & 69.09 & 20 & 71.34 & 42.03 \\
Qwen2.5~\cite{Qwen2.5-VL}  & NC & B+Mo & 14.27 & 28.91 & 30.11 & 30 & 24.73 & 25.60 \\ 
ORION~\cite{fu2025orion}  & NC & M+B+Mo & 25.00 & 71.11 & 78.33 & 30 & 69.15 & 54.72 \\
\hline
\rule{0pt}{1.2em}\ours (Qwen2.5) (ours)  & NC & B+Mo & 37.83 & 72.81 & 83.10 & 50 & 66.17 & 61.98 \\
\rule{0pt}{1.2em}\ours (ORION) (ours)  & NC & M+B+Mo & 40.06 & 78.24 & 87.32 & 50 & 76.83 & 66.49 \\
\rule{0pt}{1.2em}\ours (ORION) (ours)$\ddagger$  & TP & M+B+Mo & 48.11 & \textbf{81.16} & \textbf{88.66} & 50 & 77.75 & \textbf{69.14} \\
\hline
\end{tabular} }
\vspace{-7pt}
\end{table*}

\begin{figure*}[t!]
    \centering
    \includegraphics[width=0.99\linewidth]{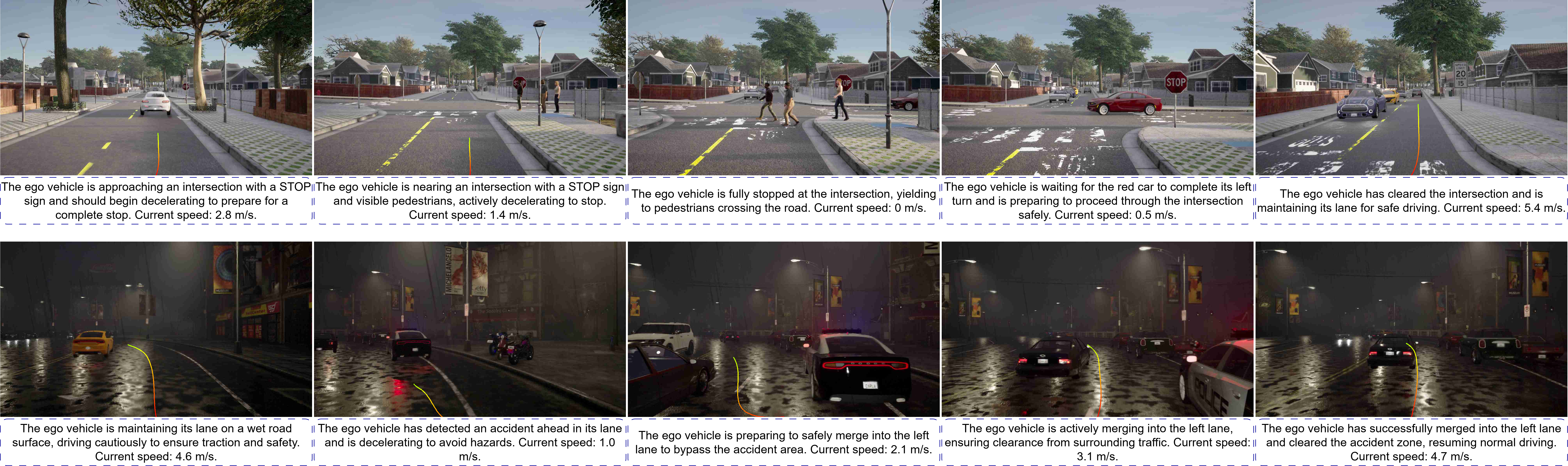}
    \vspace{-5pt}
    \caption{\small Qualitative examples of language-guided scenario rollouts using proposed \ours on Bench2Drive. \textbf{Top:} Intersection scenarios with STOP signs and pedestrian interactions, where the ego vehicle demonstrates cautious deceleration, yielding, and safe left-turn. \textbf{Bottom:} Accident-handling scenarios under adverse weather, showing proactive hazard detection, lane-change planning, and smooth merging. Each frame includes generated textual reasoning aligned with the ego actions, highlighting temporal consistency and safety-aware decision.}
    \label{fig:GeSA_qual}
    \vspace{-15pt}
\end{figure*}

\vspace{-6pt}
\subsection{Implementation Details}
\vspace{-5pt}
\paragraph{Model Architecture.} 
We implement the \ours framework on two models. \ours~(Qwen) is built on the general multi-modal LLM Qwen2.5-3B-VL~\cite{bai2023qwenvl} trained for E2E motion planning. \ours~(ORION) is built on the VLA model ORION ~\cite{fu2025orion}. In each case, \ours leverages the vision encoder, text encoder, and the LLM of Qwen2.5. The vision encoder is the EVA-pretrained vision transformer~\cite{fang2023eva}. The motion prediction head is a stack of three MLP layers. The generative planning head is a variational autoencoder (VAE) from~\cite{fu2025orion}. In \ours~(ORION), we utilize the auxiliary information and tasks used by the ORION model, while \ours~(Qwen) does not depend on HD maps. 
\ours~(Qwen) predicts trajectories conditioned only on the Navigation Command (NC), without explicit lane-center targets such as the Target Point (TP). \ours~(Qwen) also employs an anchor-free design, generating six trajectory modes aligned with those defined in Bench2Drive. Note,  autoregressive scenario rollout is used during training and disabled by default at inference; Table~\ref{tab:inference_rollout} evaluates optional inference-time rollout.

\vspace{-10pt}
\paragraph{Training.} 


All experiments are conducted on 8 NVIDIA H100 GPUs. \ours is trained for 24 epochs in Stage‑1 to learn robust latent token representations, followed by 24 epochs in Stage‑2 with autoregressive scenario rollouts of length \(T\) (\(t \!\to\! t+T\)) optimized using the proposed losses. Stage-2 consists of 20 epochs of imitation learning using the rollout loss \(\mathcal{L}_{\mathrm{roll}}\), followed by 4 epochs of reinforcement learning using the GRPO objective \(\mathcal{L}_{\mathrm{GRPO}}\). Stage‑1 training takes approximately 7 days, while Stage‑2 training takes approximately 10 days. We use \(t=4\) input frames, a rollout horizon \(T=8\), and a uniform stride \(r \in [1,4]\), covering up to 32 future frames. All Qwen2.5‑3B parameters are updated in both stages. Training uses AdamW with cosine annealing, a learning rate of \(2\times10^{-4}\), and weight decay 0.01. GRPO uses a rollout group size of \(G=8\), a minibatch size of 32, and a clipping parameter \(\epsilon=0.1\), with a frozen behavior policy and detached rewards, group baselines, and normalized advantages, as defined in Eq.~\ref{eqn:grpo-loss}.

\vspace{-6pt}
\section{Results and Discussion}
\vspace{-7pt}
We perform a thorough evaluation of \ours in both closed-loop and open-loop planning settings. We demonstrate the capability of \ours as a general plug-and-play framework for VLA models by implementing it on two end-to-end motion planners, Qwen2.5 and ORION. We compare the performance of \ours~(Qwen) and \ours~(ORION) against recent end-to-end planners as well as the baselines: 1) The original ORION VLA model as proposed in~\cite{fu2025orion}, 2) ORION (Qwen2.5) which replaces the original Vicuna language model with that of Qwen2.5, and 3) Qwen2.5 with added trajectory planning and motion prediction heads. 
\vspace{-5pt}
\subsection{Closed-Loop Evaluation on Bench2Drive}
\vspace{-5pt}
\paragraph{Quantitative Results.}
Table~\ref{tab:b2d_results} compares quantitative closed-loop performance on 220 Bench2Drive test routes. \ours achieves substantial gains over prior state‑of‑the‑art methods. The baseline Qwen2.5 model attains only 31.6\% success rate and a driving score of 63.9, whereas \ours~(Qwen), equipped with guided scenario‑generation rollouts and GRPO‑based reinforcement learning, boosts performance to 79.6 driving score (+15.7) and 57.8\% success rate (+26.2). Applying \ours to ORION further improves this baseline, raising driving score from 77.74 to 81.90 (+4.16) and success rate from 54.62\% to 60.10\% (+5.5\%). Notably, this model outperforms both baseline ORION models, which depend on HD map and bounding box annotations. 
A comparison of multi-ability performance is shown in Table~\ref{tab:multi_ability}. This table highlights similar trends: \ours~(Qwen) improves mean ability by 140\% over the baseline Qwen2.5 (\ie~25.6 $\rightarrow$ 61.98), while \ours (ORION) delivers a 21.5\% (\ie~54.72 $\rightarrow$ 66.49) improvement over ORION. Performance gains span critical skills such as merging, overtaking, emergency braking, yielding, and traffic‑sign compliance—underscoring the effectiveness of language‑guided scenario rollouts and RL for robust action grounding. Appendix~\ref{sec:cross_skill} further analyzes behavior-level transfer between emergency braking and overtaking.
\vspace{-10pt}
\paragraph{Qualitative Results.} Figure~\ref{fig:GeSA_qual} illustrates \ours’s language-conditioned rollouts in complex urban environments, encompassing challenges such as occlusions, pedestrian crossings, yielding interactions, and wet road conditions. The ego vehicle dynamically adapts its trajectory in response to evolving scene context—slowing near crosswalks, yielding to oncoming or turning vehicles, and proceeding only after potential conflicts have cleared. These behaviors align with the reasoning articulated in the generated ego-action responses. Such alignment between textual intent and physical execution underscores how scenario-grounded rollouts enable safe, context-aware planning.

\vspace{-10pt}
\paragraph{Inference-Time Rollout.}
Table~\ref{tab:inference_rollout} compares predictions with and without autoregressive rollout at inference. For \ours~(ORION), rollout improves DS/SR from 81.9/60.1 to 85.7/63.9 and reduces ego Avg.~L2 from 0.57 to 0.46 and agent minADE/minFDE from 1.88/2.08 to 1.10/1.36. The geometric-waypoint/SimLingo-controller variant also improves from 87.2/68.8 to 89.9/71.2 DS/SR. These gains come with higher latency: \ours~(ORION) increases from 0.306 to 0.940 seconds per frame. We therefore disable inference-time rollout by default, while retaining it as an optional accuracy--latency tradeoff.

\begin{table*}[t!]
\centering

\caption{\small Driving performance using inference-time autoregressive rollout on Bench2Drive. Driving Score (DS) and Success Rate (SR) measure closed-loop driving. Ego Avg.~L2 and agent minADE/minFDE measure open-loop prediction errors.  $\ddagger$ indicates using geometric path waypoint supervision and SimLingo PID controller.}
\vspace{-0.4em}
\label{tab:inference_rollout}
\small
\setlength{\tabcolsep}{4pt}
\resizebox{0.75\textwidth}{!}{
\begin{tabular}{lccccccc}
\toprule
\multirow{3}{*}{Method} & \multirow{3}{*}{Inference Rollout} & \multicolumn{2}{c}{Closed-loop} & \multicolumn{3}{c}{Open-loop} & \multirow{3}{*}{\makecell{Latency\\(second/frame)$\downarrow$}} \\
\cmidrule(lr){3-4} \cmidrule(lr){5-7}
& & DS$\uparrow$ & SR(\%)$\uparrow$ & \makecell{Ego\\Avg.~L2$\downarrow$} & \makecell{Agent\\minADE$\downarrow$} & \makecell{Agent\\minFDE$\downarrow$} & \\
\midrule
ORION~\cite{fu2025orion} & No & 76.9 & 53.9 & 0.63 & 4.16 & 4.85 & 0.292 \\
ORION~\cite{fu2025orion} & Yes & 77.7 & 54.4 & 0.61 & 3.67 & 4.17 & 0.918 \\
\midrule
\ours~(ORION) & No & 81.9 & 60.1 & 0.57 & 1.88 & 2.08 & 0.306 \\
\ours~(ORION) & Yes & 85.7 & 63.9 & 0.46 & 1.10 & 1.36 & 0.940 \\
\bottomrule

\end{tabular}
}
 \vspace{-5pt}
\end{table*}

\begin{table*}[t!]
\centering
\vspace{-0.0em}

\begin{minipage}[t]{0.64\textwidth}
\centering
\captionof{table}{\small Evaluation of E2E methods on nuScenes~\cite{caesar2020nuscenes} validation, using PARA-Drive/TOKEN setting. $\dagger$ indicates zero-shot testing trained on Bench2Drive and evaluated on nuScenes. Ego-status is not used.}
\label{tab:nusc_motion_planning}
\vspace{-0.4em}
\resizebox{\linewidth}{!}{
\begin{tabular}{ll|cccc|cccc}
\toprule
\multirow{2}{*}{Method} & \multirow{2}{*}{\makecell{Aux. Sup.}} &
\multicolumn{4}{c|}{L2 ($\downarrow$)} &
\multicolumn{4}{c}{Collision Rate ($\downarrow$)} \\
\cmidrule(lr){3-6} \cmidrule(lr){7-10}
& & 1s & 2s & 3s & Avg. & 1s & 2s & 3s & Avg. \\
\midrule
UniAD & M+B+Mo+O & 0.48 & 0.96 & 1.65 & 1.03 & 0.05 & 0.17 & 0.71 & 0.31 \\
VAD-Base & M+B+Mo & 0.54 & 1.15 & 1.98 & 1.22 & 0.04 & 0.39 & 1.17 & 0.53 \\
GenAD & M+B+Mo & 0.36 & 0.83 & 1.55 & 0.91 & 0.06 & 0.23 & 1.00 & 0.43 \\
Senna & M+B+Mo & 0.37 & 0.54 & 0.86 & 0.59 & 0.09 & 0.12 & 0.33 & 0.18 \\
Qwen2.5 & B+Mo & 0.52 & 0.80 & 1.56 & 0.96 & 0.10 & 0.37 & 1.32 & 0.60 \\
ORION & M+B+Mo & 0.43 & 0.64 & 1.01 & 0.69 & 0.09 & 0.29 & 0.94 & 0.44 \\
\hline
\ours (Qwen2.5)$\dagger$ & B+Mo & 0.24 & 0.41 & 0.70 & 0.45 & 0.11 & 0.22 & 0.42 & 0.25 \\
\ours (Qwen2.5) & B+Mo & 0.16 & 0.27 & 0.50 & 0.31 & 0.06 & 0.10 & 0.26 & 0.14 \\
\ours (ORION)$\dagger$ & M+B+Mo & 0.16 & 0.32 & 0.61 & 0.36 & 0.08 & 0.16 & 0.39 & 0.21 \\
\ours (ORION) & M+B+Mo & 0.12 & 0.24 & 0.46 & 0.27 & 0.05 & 0.09 & 0.22 & 0.12 \\
\bottomrule
\end{tabular}}
\end{minipage}
\hfill
\begin{minipage}[t]{0.34\textwidth}
\vspace{1.5em}
\centering
\captionof{table}{\small NAVSIM-v1 NavTest results. Qwen and \ours~(Qwen) use camera-only inputs. NC: No at-fault Collision; EP: Ego Progress; PDMS: Predictive Driver Model Score. Higher is better.}
\label{tab:navsim_results}
\vspace{-0.1em}
\resizebox{\linewidth}{!}{
\begin{tabular}{lccc}
\toprule
Method & NC$\uparrow$ & EP$\uparrow$ & PDMS$\uparrow$ \\
\midrule
DiffusionDrive~\cite{liao2025diffusiondrive} & 98.2 & 82.2 & 88.1 \\
Qwen & 97.7 & 73.0 & 84.9 \\
\ours~(Qwen) & 98.9 & 87.1 & 89.7 \\
\bottomrule
\end{tabular}
  }
\end{minipage}

\vspace{-0.5em}
\end{table*}

\begin{table*}[t!]
   \caption{\small Ablation study of \ours (Qwen) on Bench2Drive. Baseline: Qwen2.5-3B. Metrics: Driving Score (DS) and Success Rate (SR).}
\centering
\label{tab:ablation_results}
\vspace{-2pt}
\resizebox{0.66\linewidth}{!}{
  \begin{tabular}{l
                  *{3}{c}
                  *{3}{c}
                  *{3}{c}
                  c c
                  *{2}{c}}
    \toprule
    \multirow{3}{*}{\textbf{ID}} & \multirow{3}{*}{\makecell{\textbf{Sc }\\\textbf{Desc}}} &
      \multirow{3}{*}{\textbf{\(\mathcal{Q}\)}} &
      \multicolumn{3}{c}{\textbf{Pretraining}} &
      \multicolumn{6}{c}{\textbf{Scenario Generation Rollout}} &
      
      \multicolumn{2}{c}{\makecell{\textbf{Closed-loop}\\\textbf{metric}}} \\
    \cmidrule(lr){4-6} \cmidrule(lr){7-12} \cmidrule(lr){13-14}
     & & & $\mathcal{L}_{plan}$ & $\mathcal{L}_{\text{mot}}$ & $\mathcal{L}_{\text{VQA}}$ 
     & \multicolumn{3}{c}{$\mathcal{L}_{roll}$} 
     & \multicolumn{3}{c}{$\mathcal{L}_{\text{GRPO}}$} 
     &  \textbf{DS} & \textbf{SR} \\ \cmidrule(lr){7-9} \cmidrule(lr){10-12}
     &  & & &  &  
     & $\mathcal{L}_{\text{tc}}$ & $\mathcal{L}_{plan}$ & $\mathcal{L}_{\text{mot}}$
     & \text{Coll} & \text{TTC} & $\mathcal{L}_{\text{VLA}}$
     &   &  \\
    \midrule
    1 & \ding{55} & \ding{55} & \ding{51} & & & & & & & & & 58.2 & 27.4\\
    2 & \ding{55} & \ding{55} & \ding{51} & \ding{51} & & & & & & & & 61.1 & 28.2 \\
    3 & \ding{51} & \ding{55} & \ding{51} & \ding{51} & \ding{51} & & & & & & & 63.9& 31.6\\ \midrule
    4 & \ding{51} & \ding{55} & \ding{51} & \ding{51} & \ding{51}& \ding{51}& & & & & & 67.8& 40.1\\
    5 & \ding{51} & \ding{55} & \ding{51} & \ding{51} & \ding{51}& \ding{51}& \ding{51}& & & & & 69.2 & 46.6\\
    6 & \ding{51} & \ding{55} & \ding{51} & \ding{51} & \ding{51} & \ding{51}& \ding{51} & \ding{51} & & &  & 74.4 & 50.4\\
    7 & \ding{51} & \ding{55} & \ding{51} & \ding{51} & \ding{51} & \ding{51}& \ding{51}& \ding{51} & \ding{51} & & & 76.9 & 54.4\\
    8 & \ding{51} & \ding{55} & \ding{51} & \ding{51} & \ding{51}& \ding{51}& \ding{51} & \ding{51} & \ding{51} & \ding{51}& & 78.5 & 55.7 \\
    9 & \ding{51} & \ding{51} & \ding{51} & \ding{51} & \ding{51}& \ding{51}& \ding{51}& \ding{51} & \ding{51} & \ding{51} & \ding{51} & 79.6& 57.8\\
    \bottomrule
  \end{tabular}}
\vspace{-1em}
\end{table*}

\vspace{-5pt}
\subsection{Open-Loop Evaluation}
\vspace{-5pt}
We further evaluate open-loop trajectory prediction on the nuScenes and Bench2Drive planning benchmarks. This assesses both generalization and motion-forecasting quality. Our proposed \ours delivers substantial gains over Qwen2.5 and ORION baselines. In Table \ref{tab:b2d_results}, \ours shows a consistent drop in trajectory error across both baselines on Bench2Drive. Table \ref{tab:nusc_motion_planning} contains open-loop planning results on nuScenes. Compared to Qwen2.5, \ours (Qwen) reduces L2 trajectory error from 0.96 to 0.31 (-67.7\%) and collision rate from 0.60 to 0.14 (-76.7\%). In the zero-shot setting, errors still drop to 0.45 (-53.1\%) with collisions at 0.25 (-58.3\%), highlighting strong cross-dataset generalization enabled by scenario-grounded rollouts. Using ORION as the backbone, \ours (ORION) further improves L2 trajectory error from 0.69 to 0.27 (-60.9\%) and collision rate from 0.44 to 0.12 (-72.7\%). Zero-shot again shows notable gains (0.36, -47.8\%; 0.21, -52.3\%). Appendix~\ref{sec:qwen_controlled_ablation} details the controlled Qwen2.5 baseline and progressive training ablation.
\vspace{-10pt}
\paragraph{NAVSIM Evaluation.}
Table~\ref{tab:navsim_results} complements displacement-error evaluation with planning-oriented metrics on NAVSIM-v1 NavTest. With camera-only inputs, \ours~(Qwen) improves NC from 97.7 to 98.9, EP from 73.0 to 87.1, and PDMS from 84.9 to 89.7 over Qwen. We include DiffusionDrive~\cite{liao2025diffusiondrive} as a literature reference. These results provide additional evidence of planning gains beyond Bench2Drive on real-world driving scenes.

\subsection{Ablation Study and Further Performance Analysis}
\vspace{-7pt}
Table~\ref{tab:ablation_results} summarizes the ablation experiments of \ours on the Bench2Drive dataset using Qwen2.5-3B as the baseline model. We report the closed-loop metrics Driving Score (DS) and Success Rate (SR). Additional ablation study and performance analysis can be found in the appendix, including planner design, pretraining, rollout parameters, cross-skill evaluation, and rollout realism.

\vspace{-11pt}
\paragraph{Pretraining.} IDs 1–3 in Table~\ref{tab:ablation_results} show the impact of pretraining losses \(\mathcal{L}_{\text{plan}}, \mathcal{L}_{\text{mot}}, \mathcal{L}_{\text{VLA}}\). Incorporating scenario descriptions and visual question answering (VQA) improves baseline performance, indicating that language grounding during pretraining enhances planning robustness.

\vspace{-11pt}
\paragraph{Scenario Rollout.} IDs 4–6 evaluate contributions of rollout-consistency losses \(\mathcal{L}_{\text{roll}}\) within the proposed \ours  framework. Even under model-based supervision—where ground truth is not required and we rely on predicted tokens \(\{z_{t+1}\}\) from the pretrained VLA using input image sequence \(I_{1:t}\) as reference targets—computing \(\mathcal{L}_{\text{tc}}\) yields notable gains (+3.9 DS, +8.5\% SR). Further incorporating ground-truth trajectories \(\{\tau_{t+\Delta}\}\) for \(\mathcal{L}_{\text{mot}}\) and \(\mathcal{L}_{\text{plan}}\) boosts performance by an additional +6.6 DS and +10.3\% SR, demonstrating the value of accurate trajectory supervision during rollouts.

\vspace{-11pt}
\paragraph{RL with GRPO.} IDs 7–9 analyze the effect of GRPO-based reinforcement learning losses \(\mathcal{L}_{\text{GRPO}}\). Using collision and time-to-collision (TTC) rewards during scenario rollouts improves DS by +4.1 and SR by +5.3\%. Finally, introducing ego-action question–answer pairs \(\mathcal{Q}\) (Figure~\ref{fig:GeSA_rollout}) and leveraging the corresponding language alignment loss \(\mathcal{L}_{\text{VLA}}\) as an additional reward further enhances performance (+1.1 DS, +2.1\% SR), highlighting the benefit of grounding ego intent in multi-step maneuvers through language-conditioned rollouts.

\section{Conclusion}\label{sec:conclusion}
\vspace{-7pt}
In this work, we introduced \ours, a generative scenario rollout framework that unifies planning, multi-agent motion prediction, and scenario generation for robust end-to-end planning with Vision-Language-Action models. \ours leverages shared latent tokenization and language-conditioned autoregressive rollout to enable temporally consistent and context-aware driving behavior. By integrating reinforcement learning with GRPO and aligning textual reasoning with action execution, our proposed \ours sets the new state-of-the-art performance with its robust, language-guided end-to-end planning. On standard autonomous driving benchmarks, \ours provides significant improvements under both closed-loop and open-loop settings, as well as strong zero-shot sim-to-real generalization. 
\vspace{-7pt}
\paragraph{Broader Impact.}
\label{sec:impact_statement}
This work introduces Generative Scenario Rollouts (GeRo), a plug-and-play training framework for Vision-Language-Action (VLA) models that jointly performs planning and language-grounded future scenario generation via autoregressive rollouts in a learned latent token space.  By enabling temporally consistent language-conditioned rollouts (and an interactive VQA component that verbalizes ego intent), GeRo can improve long-horizon reasoning and interpretability for end-to-end autonomous driving policies, pushing the boundaries of autonomous driving.
\vspace{-7pt}
\paragraph{Limitations}
\label{sec:limitations}
Although GeRo jointly predicts ego plans and multi-agent futures, it fundamentally plans only for the ego vehicle and treats other agents as forecasts rather than decision-making entities, which can limit fidelity in highly interactive scenarios (e.g., adversarial behaviors of other agents) where other agents react strategically.







{
    \small
    \bibliographystyle{ieeenat_fullname}
    \bibliography{main}
}

\appendix

\clearpage


\section{Ablation Study}
\subsection{Ablation on Generative Planners}
Table~\ref{tab:gen_planner} presents an ablation study comparing different generative planners on Bench2Drive dataset under the closed-loop setting. The reported metrics include driving score (DS) and success rate (SR) for closed-loop performance, and average L2 trajectory error for open-loop trajectory prediction. We also provide an overall multi-ability average to summarize planner effectiveness. The results demonstrate that incorporating generative scenario rollouts significantly improves performance across all metrics. Specifically, our proposed approach (\ours) consistently outperforms the baseline across different variants of generative planners architecture (MLP, Diffusion, or VAE). Among these, the VAE-based planner achieves the highest scores, with \ours (ORION) w/ VAE reaching a Driving Score of 81.90 and a Success Rate of 60.10\%, while reducing the L2 error to 0.57\,m. These findings highlight the advantage of combining generative scenario rollouts with diverse planning strategies.

\begin{table*}[h!]
    \centering
    \caption{Ablation comparison with diverse planners. DS and SR denote Driving Score and Success Rate respectively.}
    \label{tab:gen_planner}
    \vspace{-6pt}
    \resizebox{0.6\textwidth}{!}{
    \begin{tabular}{lcccc}
        \toprule
        \multirow{2}{*}{Generative Planner} & \multicolumn{2}{c}{Closed-loop} & \multicolumn{1}{c}{Open-loop} & \multirow{2}{*}{Multi-Ability Avg.} \\
        \cmidrule(lr){2-3} \cmidrule(lr){4-4}
        & DS$\uparrow$ & SR(\%)$\uparrow$ & Avg. L2 (m)$\downarrow$ & \\
        \midrule
        ORION w/ MLP & 70.73 & 41.52 & 0.75 & 48.44 \\
        ORION w/ Diffusion & 71.97 & 46.54 & 0.73 & 46.68 \\
        ORION w/ VAE & 77.74 & 54.62 & 0.68 & 54.72 \\ \midrule
        \ours (ORION) w/ MLP & 77.21 & 56.72 & 0.62 & 62.38 \\
        \ours (ORION) w/ Diffusion & 80.75 & 59.22 & 0.57  & 65.43 \\
        \ours (ORION) w/ VAE & 81.90 & 60.10 & 0.57 & 66.49\\
        \bottomrule
    \end{tabular}
    }
\end{table*}

\subsection{Ablation on Stage-1 Pretraining}
Table~\ref{tab:gen_stage1} reports an ablation study evaluating the effect of Stage-1 pretraining on our generative planner based on Qwen2.5.  For this experiment, we use Qwen2.5-3B as the baseline for \ours. We measure closed-loop performance using driving score (DS) and success rate (SR), along with a multi-ability average summarizing overall capability. The baseline Qwen2.5 achieves a DS of 63.90 and SR of 31.60\%, indicating limited performance without generative rollouts. Incorporating our approach (\ours) without Stage-1 pretraining significantly improves results to 71.67 DS and 48.95\% SR, demonstrating the benefit of scenario rollouts alone. Further applying Stage-1 pretraining yields the best performance, achieving 79.60 DS and 57.80\% SR, with a Multi-Ability Average of 61.98. These results highlight the significance of stage-1 pretraining in scenario rollout generation framework of \ours 
\begin{table*}[h!]
    \centering
    \caption{Ablation comparison on stage-1 pretraning. DS and SR denote Driving Score and Success Rate respectively.}
    \label{tab:gen_stage1}
    \vspace{-6pt}
    \resizebox{0.6\textwidth}{!}{
    \begin{tabular}{lccc}
        \toprule
        \multirow{2}{*}{Generative Planner} & \multicolumn{2}{c}{Closed-loop} & \multirow{2}{*}{Multi-Ability Avg.} \\
        \cmidrule(lr){2-3}
        & DS$\uparrow$ & SR(\%)$\uparrow$ & \\
        \midrule
        Qwen2.5 & 63.90 & 31.60 & 25.60\\
        \ours (Qwen2.5) w/o stage-1 pretraining & 71.67 & 48.95 & 53.7\\
        \ours (Qwen2.5) w/ stage-1 pretraining & 79.60 & 57.80 & 61.98\\
        
        \bottomrule
    \end{tabular}
    }
\end{table*}

\subsection{Controlled Qwen2.5 Baseline and Training Ablation}
\label{sec:qwen_controlled_ablation}
\ours~(Qwen2.5) uses the same backbone, tokenizer, decoder, inputs, and evaluator as the Qwen2.5 baseline, adding the proposed rollout-based training objectives. After Stage-1 pretraining, Stage-2 introduces rollout supervision for temporal consistency, agent motion, and ego planning, followed by GRPO-based safety feedback. The comparison thus isolates the effect of rollout training with the architecture, inputs, training data, and evaluation protocol held fixed. The baseline Avg.~L2 of 0.96 is specific to this controlled setup; we do not claim that it is the best attainable Qwen2.5 result across protocols.

Table~\ref{tab:nusc_training_ablation} reports the progressive ablation on nuScenes validation under the PARA-Drive/TOKEN evaluation protocol. Starting from Qwen2.5 trained with $\mathcal{L}_{\text{pre}}$, cumulatively adding temporal consistency, motion rollout, planning rollout, and GRPO feedback reduces Avg.~L2 from 0.96 to 0.31 and collision rate from 0.60 to 0.14. Each additional component improves both metrics. Table~\ref{tab:ablation_results} in the main paper provides the complementary Bench2Drive ablation using closed-loop metrics and the Qwen2.5 baseline.

\begin{table}[h!]
\centering
\caption{Training ablation of \ours~(Qwen2.5) on nuScenes validation under the PARA-Drive/TOKEN evaluation protocol. Baseline: Qwen2.5-3B-VL. Objectives are added cumulatively; lower is better.}
\label{tab:nusc_training_ablation}
\small
\begin{tabular}{llcc}
\toprule
Stage & Model / added objective & Avg.~L2$\downarrow$ & Collision rate$\downarrow$ \\
\midrule
$\mathcal{L}_{\text{pre}}$ & Qwen2.5 & 0.96 & 0.60 \\
Stage-2 $\mathcal{L}_{\text{roll}}$ & $+\mathcal{L}_{\text{tc}}$ & 0.61 & 0.31 \\
Stage-2 $\mathcal{L}_{\text{roll}}$ & $+\mathcal{L}_{\text{mot}}$ & 0.54 & 0.28 \\
Stage-2 $\mathcal{L}_{\text{roll}}$ & $+\mathcal{L}_{\text{plan}}$ & 0.39 & 0.20 \\
Stage-2 $\mathcal{L}_{\text{GRPO}}$ & $+\mathcal{L}_{\text{GRPO}}$ & 0.31 & 0.14 \\
\bottomrule
\end{tabular}
\end{table}

\subsection{Ablation on $T$, Rollout Steps in Scenario Generation}
Table~\ref{tab:ablation_T} shows an ablation study on the number of rollout steps ($T$) used during scenario generation. For this experiment, we use Qwen2.5-3B as the baseline for \ours. We report closed-loop performance using driving score (DS) and success rate (SR). Increasing $T$ from 4 to 12 leads to consistent improvements, with DS rising from 78.31 to 80.15 and SR from 56.75\% to 58.01\%. This indicates that longer rollouts provide richer scenario context, improving planning quality. However, performance slightly drops at $T=16$, suggesting diminishing returns and potential overfitting or increased complexity at very high rollout lengths.
\begin{table}[h!]
\centering
\caption{Ablation on $T$, the number of rollout steps in scenario generation.}
\label{tab:ablation_T}
\resizebox{0.5\columnwidth}{!}{
\begin{tabular}{lcccc}
\toprule
Metric & $T=4$ & $T=8$ & $T=12$ & $T=16$ \\
\midrule
Driving Score$\uparrow$ & 78.31 & 79.60 & 80.15 & 79.08 \\
Success Rate$\uparrow$ & 56.75 & 57.80 & 58.01 & 57.17 \\
\bottomrule
\end{tabular}
}
\end{table}

\subsection{Behavior Diversity and Cross-Skill Generalization}
\label{sec:cross_skill}
Future driving behavior can be multi-modal. \ours encourages behavior-level diversity through language-conditioned rollout training, but is not a full probabilistic simulator covering all possible agent futures. The multi-ability results in Table~\ref{tab:multi_ability} show improvements on multiple driving skills, including an increase in the mean score from 54.72 for ORION to 66.49 for \ours~(ORION). We further assess transfer between behavior modes using skill-specific training subsets.

Bench2Drive contains 44 scenarios grouped into five advanced urban-driving skills: Merging, Overtaking, Give Way, Traffic Sign, and Emergency Brake. We train models using only Emergency Brake routes or only Overtaking routes, and evaluate each model on both skills using the Bench2Drive multi-ability metric. During Stage-2 training of \ours, we counterfactually perturb ego-action descriptions to alternative high-level behaviors, such as braking or overtaking. The model generates ego waypoints and agent trajectories conditioned on the perturbed description, while GRPO safety feedback encourages collision avoidance and high time-to-collision (TTC).

\begin{center}
\begin{minipage}{\linewidth}
\centering
\captionof{table}{Cross-skill evaluation on Bench2Drive. Scores use the multi-ability metric (\%); higher is better. Full-training results are included for reference.}
\label{tab:cross_skill}
\small
\setlength{\tabcolsep}{5pt}
\begin{tabular}{llcc}
\toprule
Model & Training routes & \makecell{Emergency\\Brake$\uparrow$} & Overtaking$\uparrow$ \\
\midrule
Qwen & Emergency Brake only & 39.81 & 10.44 \\
\ours~(Qwen) & Emergency Brake only & 90.38 & 70.15 \\
\midrule
Qwen & Overtaking only & 9.75 & 34.16 \\
\ours~(Qwen) & Overtaking only & 81.86 & 84.21 \\
\midrule
Qwen & Full Bench2Drive training & 30.11 & 28.91 \\
\ours~(Qwen) & Full Bench2Drive training & 83.10 & 72.81 \\
\bottomrule
\end{tabular}
\end{minipage}
\end{center}

Table~\ref{tab:cross_skill} shows substantial transfer gains: with Emergency Brake-only training, the Overtaking score improves from 10.44 for Qwen to 70.15 for \ours~(Qwen); with Overtaking-only training, the Emergency Brake score improves from 9.75 to 81.86. These results support behavior-level generalization across training subsets. They do not directly quantify the diversity or calibration of multiple plausible futures for a fixed scene, or isolate how such future diversity affects ego planning; these remain limitations of the analysis.

\subsection{GPT Score, CIDEr, and BLEU Results}
Table~\ref{tab:gpt_score} presents the quality of scenario generation and ego-action question-answer (QA) pairs produced by \ours~ during scenario generation rollouts. For this experiment, we use Qwen2.5-3B as the baseline for \ours. We report quality of text generation or language outputs of \ours using GPT score~\cite{fu2023gptscore}, CIDEr~\cite{vedantam2015cider}, and BLEU~\cite{papineni2002bleu}. Our method demonstrates substantial improvements over the baseline across all language evaluation metrics on the Bench2Drive benchmark. For the Ego-Action task, GPT score increased from 70.82 to 77.54, indicating the effectiveness of language-grounded scenario rollouts in the \ours~ framework. Similarly, CIDEr and BLEU scores improved by 7.31 and 11.69 points, respectively, reflecting stronger alignment with ego-planning actions. For Scenario Description, our approach achieved an 8.31-point gain in GPT score and a 10.80-point increase in CIDEr, confirming its ability to generate richer and more accurate scenario narratives. These consistent improvements across both tasks highlight the robustness of our method in autonomous driving.
\begin{table}[h!]
\centering
\caption{Language performance on Bench2Drive for ego-action QA and scenario description.}
\label{tab:gpt_score}
\resizebox{0.7\columnwidth}{!}{
\begin{tabular}{l ccc cc}
\toprule
\multirow{2}{*}{Method} & \multicolumn{3}{c}{Ego-Action} & \multicolumn{2}{c}{Scenario Description} \\ \cmidrule(lr){2-4} \cmidrule(lr){5-6}
 & GPT score$\uparrow$  & CIDEr$\uparrow$  & BLEU$\uparrow$  & GPT score$\uparrow$  & CIDEr$\uparrow$ \\
\midrule
Qwen2.5 & 70.82 & 67.90 & 53.69 & 68.16 & 72.21 \\
\ours (Qwen2.5) & 77.54 & 75.21 & 65.38 & 76.47 & 83.01\\
\bottomrule
\end{tabular}
}
\end{table}



\begin{figure*}
    \centering

\begin{tcolorbox}[colback=gray!10, colframe=gray!50, boxrule=0.5mm, rounded corners, title={\textbf{\textcolor{black}{Scenario Description and Ego-Action Question Answer pairs  \textbar{} Sample numerical annotation}}},fontupper=\small]

\begin{verbatim}
"token": "f9878012c3f6412184c294c13ba4bac3",
"Senario description": " Pedestrain crossing at stop sign in suburban
                         street with wet road conditions",
"Critical Object information": "Pedestrian at <-2.31, 4.41> is crossing the
                                street from left to right, Pedestrian at <1.24,
                                3.83> is standing on the sidewalk. Traffic 
                                signs at various distances ahead provide 
                                speed limits."
"agent_attributes": {
    "c1": {
        "category": "pedestrian",
        "speed": 1.2 meters/second,
        "position": <-2.31, 4.41>,
        "assigned_cameras": [
            "CAM_FRONT",
            "CAM_FRONT_LEFT"
        ],
        "future movement": "moving",
        "future speed": "moving",
        "direction": "towards the ego vehicle",
        "interaction_with_ego_vehicle_type": "none",
        "probability_of_collision_with_ego_vehicle": "low"
    },
    
    ... [TRUNCATED] ...
    
    "c24": {
        "category": "car",
        "speed": 0.0,
        "assigned_cameras": [
            "CAM_BACK"
        ],
        "future movement": "stopped",
        "future speed": "not moving",
        "direction": "away from ego vehicle",
        "interaction_with_ego_vehicle_type": "none",
        "probability_of_collision_with_ego_vehicle": "low"
    }

\end{verbatim}

\end{tcolorbox}
    \caption{An example of the text-description of the numerical annotations of a scene from Bench2Drive. This is appended to the user prompt for Qwen3-VL-32B-VL.}
    \label{fig:user_prompt}
\end{figure*}

\begin{figure*}
\centering

\begin{tcolorbox}[colback=gray!10, colframe=gray!50, boxrule=0.5mm, rounded corners, title={\textbf{\textcolor{black}{Generated Scenario Description and Ego-Action Question Answer pairs with  Qwen3-VL-32B-VL }}}]

\begin{tcolorbox}[colback=orange!10, colframe=orange!50, boxrule=0.5mm, rounded corners, title={\textbf{\textcolor{black}{Scenario Description}}}, fontupper=\small]

\begin{verbatim}
    "Scenario Description":  " The ego vehicle approaches an urban 
    intersection under heavy fog and wet road conditions, preparing 
    to cross while yielding to pedestrians and oncoming traffic. 
    After detecting a fast-approaching vehicle, it pauses 
    mid-intersection before safely completing the maneuver 
    and maintaining lane stability."
\end{verbatim}
\end{tcolorbox}

\begin{tcolorbox}[colback=blue!10, colframe=blue!50, boxrule=0.5mm, rounded corners, title={\textbf{\textcolor{black}{Ego-Action question answer pair}}},fontupper=\small]

\begin{verbatim}
{
    "Question at step1": "What are the predicted future movements of the 
    agents located to the front-left of the ego vehicle at time step 1? 
    What are the predicted future movements of the ego vehicle 
    at time step 1?",
    
    "Answer at step1": "The front-left vehicle is expected to remain 
    stationary. The ego vehicle is fully stopped at the intersection, 
    yielding to pedestrians crossing the road."
},
... [TRUNCATED] ...
{
    "Question at step8": "What are the predicted future movements of the
    ego vehicle at time step 8?",
    
    "Answer at step8": "The ego vehicle has resumed motion, entered the 
    intersection, and is proceeding straight ahead, safely crossing."
},

\end{verbatim}
\end{tcolorbox}

\end{tcolorbox}
    \caption{Some examples of generated QA pairs. The perception, prediction, and planning pairs are generated with Llama-3. The behavior QA is created using the future motion of the ego vehicle.}
    \label{fig:gen_scenario}
\end{figure*}

\section{Language Annotation Example}\label{sec:language_anno}
Consider a driving scene at time $t$ consisting of an ego vehicle and a set of $N_t$ traffic participants $\mathcal{A}_t$. Given a sequence of multi-view images, a scene description, critical object information (similar to ChatB2D~\cite{fu2025orion}), and information about the state of the ego vehicle $x_t^e$ and the states of other traffic participants $x_t^{(a_i)}$, the objective is to generate a scenario description and ego-action question-answer pairs ($\mathcal{Q}$) using Qwen3-VL-32B-VL. An example scenario description and ego-action QA pair are shown in Figure~\ref{fig:user_prompt} and Figure~\ref{fig:gen_scenario}.

\newpage
\section{Scenarios}
 Bench2Drive contains 44 scenarios summarized  into 5 advanced skills for urban driving: Merging, Overtaking, Give Way, Traffic
Sign, Emergency Brake as below:

\begin{enumerate}
    \item \textbf{Merging}
    \begin{itemize}
        \item CrossingBicycleFlow
        \item EnterActorFlow
        \item HighwayExit
        \item InterurbanActorFlow
        \item HighwayCutIn
        \item InterurbanAdvancedActorFlow
        \item MergerIntoSlowTrafficV2
        \item MergeIntoSlowTraffic
        \item NonSignalizedJunctionLeftTurn
        \item NonSignalizedJunctionRightTurn
        \item NonSignalizedJunctionLeftTurnEnterFlow
        \item ParkingExit
        \item LaneChange
        \item SignalizedJunctionLeftTurn
        \item SignalizedJunctionRightTurn
        \item SignalizedJunctionLeftTurnEnterFlow
    \end{itemize}

    \item \textbf{Overtaking}
    \begin{itemize}
        \item Accident
        \item AccidentTwoWays
        \item ConstructionObstacle
        \item ConstructionObstacleTwoWays
        \item HazardAtSideLaneTwoWays
        \item HazardAtSideLane
        \item ParkedObstacleTwoWays
        \item ParkedObstacle
        \item VehicleOpenDoorTwoWays
    \end{itemize}

    \item \textbf{Emergency Brake}
    \begin{itemize}
        \item BlockedIntersection
        \item DynamicObjectCrossing
        \item HardBrakeRoute
        \item OppositeVehicleTakingPriority
        \item OppositeVehicleRunningRedLight
        \item ParkingCutIn
        \item PedestrianCrossing
        \item ParkingCrossingPedestrian
        \item StaticCutIn
        \item VehicleTurningRoute
        \item VehicleTurningRoutePedestrian
        \item ControlLoss
    \end{itemize}

    \item \textbf{Give Way}
    \begin{itemize}
        \item InvadingTurn
        \item YieldToEmergencyVehicle
    \end{itemize}

    \item \textbf{Traffic Sign}
    \begin{itemize}
        \item EnterActorFlow
        \item CrossingBicycleFlow
        \item NonSignalizedJunctionLeftTurn
        \item NonSignalizedJunctionRightTurn
        \item NonSignalizedJunctionLeftTurnEnterFlow
        \item OppositeVehicleTakingPriority
        \item OppositeVehicleRunningRedLight
        \item PedestrianCrossing
        \item SignalizedJunctionLeftTurn
        \item SignalizedJunctionRightTurn
        \item SignalizedJunctionLeftTurnEnterFlow
        \item Tjunction
        \item VanillaNonSignalizedTurnEncounterGiveWay
        \item VanillaSignalizedTurnEncounterGiveWay
        \item VanillaSignalizedTurnEncounterStopSign
        \item VanillaNonSignalizedTurnEncounterStopSign
        \item VehicleTurningRoute
        \item VehicleTurningRoutePedestrian
    \end{itemize}
\end{enumerate}

\newpage
\begin{figure*}[ht]
    \centering
    \includegraphics[width=0.99\linewidth]{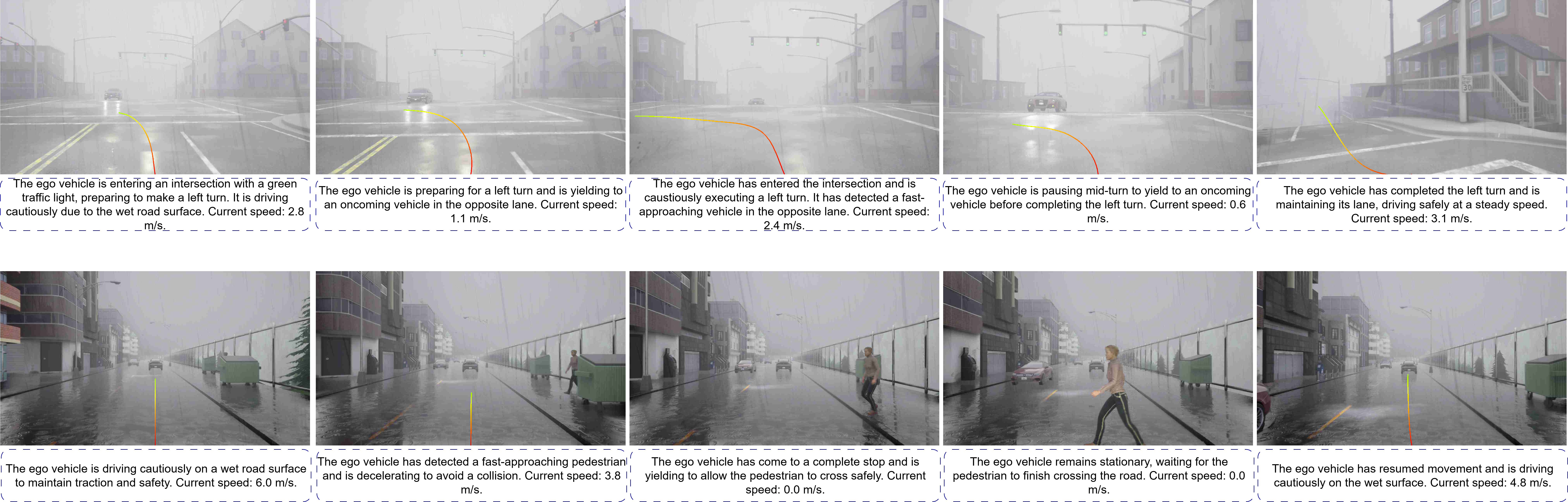}
    \vspace{-5pt}
    \caption{Qualitative visualization of two complex urban driving scenarios under adverse weather conditions. \textbf{Top:} illustrates a left-turn maneuver at an intersection with limited visibility, requiring cautious execution and yielding to oncoming traffic. \textbf{Bottom:} depicts a pedestrian-crossing scenario where the ego vehicle detects a fast-approaching pedestrian, performs an emergency stop, and resumes motion safely after the pedestrian clears the road. These examples highlight the \ours's ability to handle dynamic and safety-critical situations.}
    \label{fig:GeSA_complex}
    \vspace{-8pt}
\end{figure*}

\begin{figure*}[ht]
    \centering
    \includegraphics[width=0.90\linewidth]{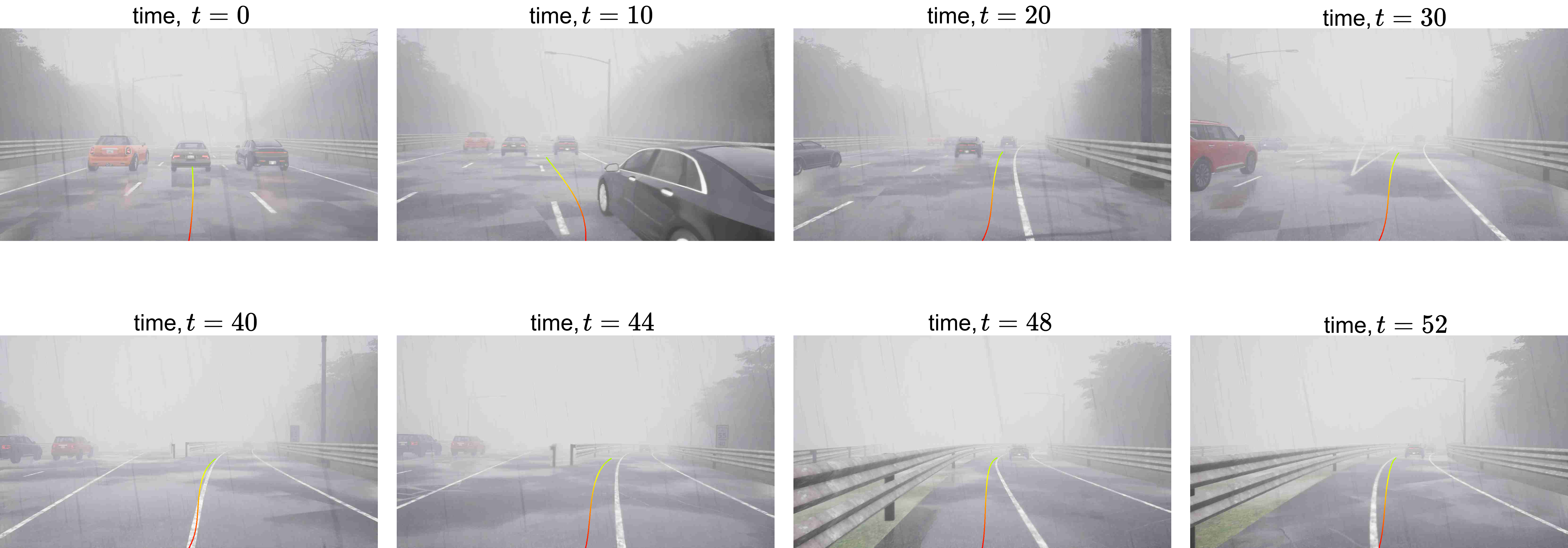}
    \caption{Failure case under extreme fog conditions. The ego vehicle initially performs a successful lane change to prepare for a highway exit (top row). However, due to low visibility, it confuses the shoulder lane with the exit lane and deviates from its route for approximately 23.476 meters (bottom row). The vehicle later corrects its trajectory and merges back into the exit lane.}
    \label{fig:GeSA_failure}
    \vspace{-8pt}
\end{figure*}

\section{Visualizations}
\subsection{Qualitative Results}
Figure~\ref{fig:GeSA_complex} illustrates two challenging urban driving scenarios under adverse weather conditions, highlighting the ability of \ours to adapt to dynamic and safety-critical events. In the first scenario (top row), the ego vehicle approaches an intersection with limited visibility and executes a cautious left turn while yielding to oncoming traffic. The trajectory shows smooth adjustments to maintain safety on a wet surface. In the second scenario (bottom row), the ego vehicle detects a fast-approaching pedestrian, performs an emergency stop, and remains stationary until the pedestrian clears the road before resuming motion. These examples demonstrate robust decision-making, safe maneuver execution, and compliance with traffic rules in unpredictable environments.
\vspace{-3pt}

\paragraph{Qualitative Analysis of Hard Brake Route.}
Figure~\ref{fig:hard_brake_gif} illustrates the qualitative output of \ours in the ``Hard Brake Route'' scenario. In this situation, surrounding traffic halts unexpectedly, requiring the ego vehicle to perform a controlled stop. The shortened trajectory indicates the ability of \ours to decelerate smoothly and maintain safety. Once traffic resumes, the ego vehicle accelerates without abrupt maneuvers, following the planned trajectory. This behavior demonstrates the generative capability of \ours to handle stop-and-go conditions effectively, ensuring both safety and comfort in dynamic environments.
\vspace{-3pt}

\subsection{Failure Case}
Figure~\ref{fig:GeSA_failure} illustrates a failure case of \ours, where the ego vehicle deviated from its route lanes for approximately 23.476 meters. The top row of Figure~\ref{fig:GeSA_failure} shows the ego vehicle successfully changing from the center lane to the rightmost lane of the highway, preparing to exit in the subsequent frames. In contrast, the bottom row depicts the impact of extreme fog and very low visibility, which caused the ego vehicle to confuse the shoulder lane with the highway exit lane. As a result, the vehicle deviated and traveled outside its designated lane for about 23.476 meters. After continuing for a short distance on the shoulder lane and encountering a roadside railing, the ego vehicle recognized the deviation and gradually merged back into the highway exit lane, resuming safe driving.

\begin{figure}[h!]
    \centering
    \animategraphics[loop,controls,width=0.45\textwidth]{1}{figures/24604/}{0000}{0040}
    \caption{Visualization of the “Hard Brake Route” scenario.}
    \label{fig:hard_brake_gif}
\end{figure}
\section{Inference Details}
\label{sec:inference_details}

Table~\ref{tab:latency} reports the per-frame inference latency of \ours measured on a single NVIDIA H100 (80GB) GPU with batch size 1. We use 10 warmup iterations followed by 50 timed iterations, and call \texttt{torch.cuda.synchronize()} before and after each timed iteration to ensure accurate GPU timing.

\begin{table}[h!]
\vspace{-8pt}
\centering
\caption{\textbf{Per-frame inference cost on Bench2Drive (validation).} Time is measured in seconds per frame; compute is reported in TFLOPs.}
\label{tab:latency}

\setlength{\tabcolsep}{6pt}
\renewcommand{\arraystretch}{1.1}

\resizebox{0.4\linewidth}{!}{%
\begin{tabular}{l cc}
\toprule
\multirow{2}{*}{\textbf{Method}} & \multicolumn{2}{c}{\textbf{Inference}} \\
\cmidrule(lr){2-3}
 & \textbf{Time (s)} $\downarrow$ & \textbf{TFLOPs} $\downarrow$ \\
\midrule
ORION          & 0.292 & 0.796 \\
GeRo (ORION)   & 0.306 & 0.798 \\
\bottomrule
\end{tabular}%
}
\vspace{-10pt}
\end{table}

\clearpage

\section{Rollout Realism}
\label{app:rollout_realism}
Closed-loop driving scores measure the deployed policy's performance, but do not directly assess the fidelity of the latent futures used during training. We therefore evaluate decoded rollout trajectories using metrics adapted from the Waymo Open Sim Agents Challenge~\citep{montali2023simagents}, with ground-truth Bench2Drive trajectories as the reference distribution. All compared checkpoints are evaluated under the same protocol using the best checkpoint from each method. DeepSight~\citep{zhang2026deepsight} provides a relevant latent world-modeling comparison.

The kinematic metric averages likelihoods of linear speed, linear acceleration, angular speed, and angular acceleration, assessing agreement with reference motion statistics. The interactive metric averages likelihoods of distance to the nearest object, collision indication, and time-to-collision, assessing proximity and collision-risk statistics. The realism meta metric is a linear aggregation of these measures. We omit the map metric because distance-to-center values are unavailable for some methods. Thus, the evaluation measures trajectory-distribution agreement rather than latent-token reconstruction or exact matching of a single future.

\begin{table}[h!]
\centering
\caption{Latent rollout realism on Bench2Drive using adapted Waymo Sim Agents metrics. Higher values indicate better reference-distribution agreement.}
\label{tab:rollout_realism}
\small
\setlength{\tabcolsep}{8pt}
\begin{tabular}{lccc}
\toprule
Method & Meta$\uparrow$ & Kinematic$\uparrow$ & Interactive$\uparrow$ \\
\midrule
ORION & 0.391 & 0.263 & 0.518 \\
SimLingo & 0.421 & 0.368 & 0.573 \\
DeepSight~\citep{zhang2026deepsight} & 0.490 & 0.413 & 0.608 \\
\ours & 0.553 & 0.454 & 0.620 \\
\bottomrule
\end{tabular}
\end{table}

Table~\ref{tab:rollout_realism} shows that \ours improves all three metrics over the compared baselines, reaching a kinematic score of 0.454, an interactive score of 0.620, and a realism meta score of 0.553, compared with 0.391 for ORION. The interactive improvement is relevant because predicted surrounding-agent trajectories shape the collision and time-to-collision feedback used during rollout-based training. These results complement the driving evaluation by assessing agreement with reference motion and interaction statistics. They do not guarantee exact physical dynamics for every generated future or exclude exploitation of imperfect rewards.
\clearpage


\end{document}